\documentclass{article}

\usepackage[preprint]{corl_2026}

\usepackage{tikz}
\usepackage{amsmath,amssymb}
\usetikzlibrary{arrows.meta,positioning,fit,calc,shapes.geometric,shapes.symbols,backgrounds}

\usepackage{float}

\usepackage{booktabs}
\usepackage{graphicx}

\usepackage{tikz}
\usetikzlibrary{arrows.meta, positioning, calc, fit}
\usepackage{subcaption}
\usepackage{amsfonts,amsmath,amssymb}
\usepackage{graphicx}
\usepackage{booktabs}
\usepackage{xcolor}
\usepackage{subcaption}
\usepackage{multirow}
\usepackage{enumitem}
\usepackage{algorithm}
\usepackage{algorithmic}

\title{Detecting is Easy, Adapting is Hard: Local Expert Growth for Visual Model-Based Reinforcement Learning under Distribution Shift}

\author{
  Haiyang Zhao\\
  University of Georgia \\
  United States\\
  \texttt{hz33711@uga.edu} \\
}

\begin{document}
\maketitle


\begin{abstract}
Visual model-based reinforcement learning (MBRL) agents can perform well on the training distribution, but often break down once the test environment shifts. A natural assumption is that, if the agent can detect such a shift, then simple remedies such as uncertainty penalties or direct policy adaptation should be enough. Our results suggest that this view is incomplete. In visual MBRL, recognizing that a shift has occurred is often the easier part; the harder part is turning that recognition into useful action-level correction. We study several ways of responding to shift, including planning penalties, direct fine-tuning, global residual correction, and coarse gating. In our experiments, these approaches either do not improve closed-loop control or hurt in-distribution (ID) performance. Based on these negative results, we propose \emph{JEPA-Indexed Local Expert Growth}. The method uses a frozen JEPA representation only for problem indexing, while cluster-specific residual experts add local action corrections on top of the original controller. The baseline controller itself is not modified. Using paired-bootstrap evaluation, we find that the original naive-preference variant is not stable under stricter testing. In contrast, the harder-pair variant produces statistically significant OOD improvements on all four evaluated shift conditions while preserving ID performance. The learned experts also remain useful when the same shift is encountered again, which supports the view of adaptation as incremental knowledge growth rather than repeated full retraining. We further show that automatic ID rejection can be achieved with simple density models, whereas fine-grained discrimination among OOD sub-families is limited by the representation. This supports our hybrid design that combines manual routing with automatic ID rejection. Overall, the results indicate that, for visual MBRL under distribution shift, the main challenge is not simply noticing that the environment has changed, but applying the right local action correction after the change has been recognized.
\end{abstract}

\keywords{Model-Based Reinforcement Learning, Distribution Shift, Joint Embedding, OOD Detection}



\begin{figure*}[t]
\centering
\includegraphics[width=\textwidth]{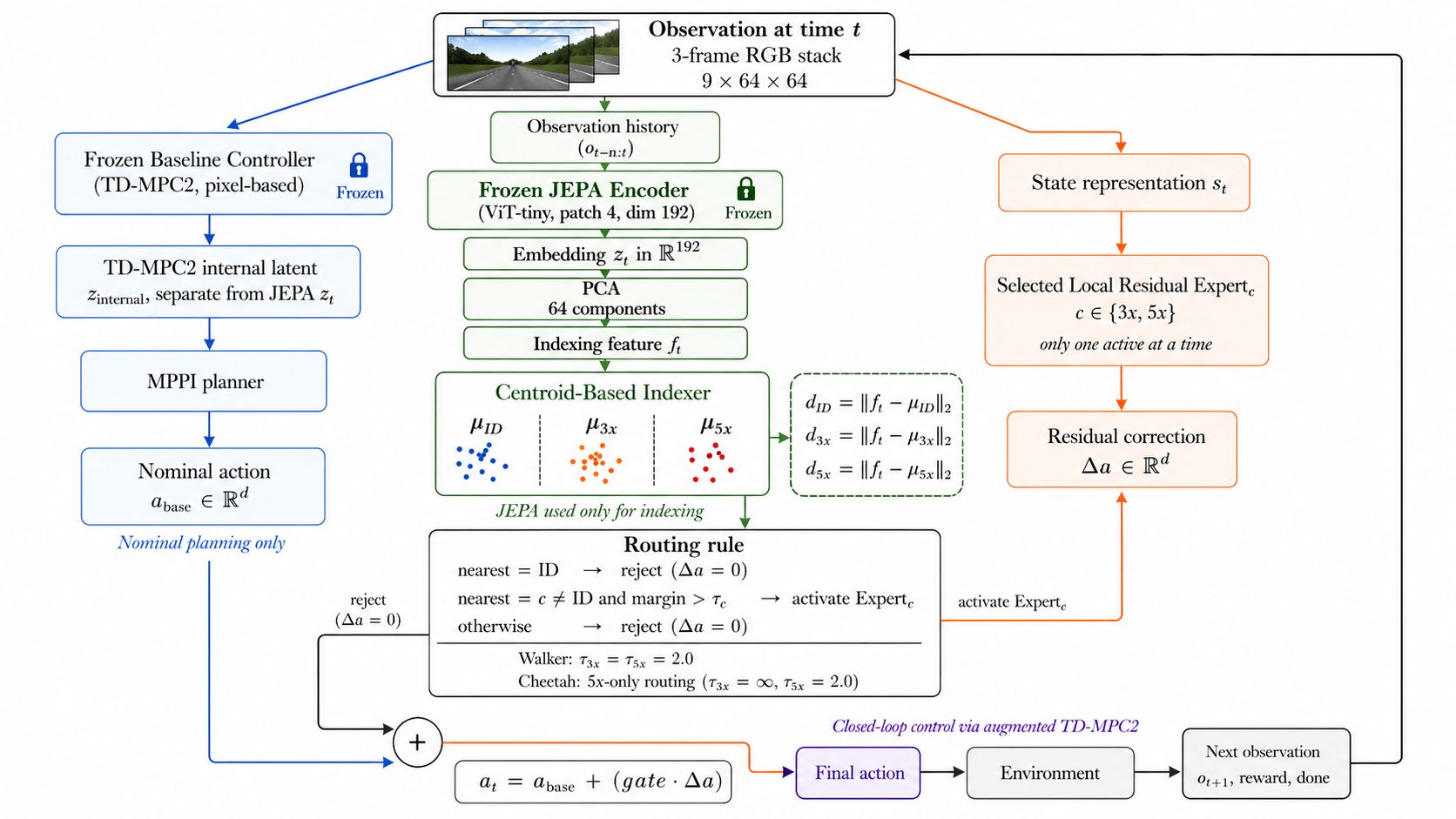}
\caption{\textbf{JEPA-Indexed Local Expert Growth.}
The baseline pixel controller produces a base action $a_t^{\mathrm{base}}$. A frozen JEPA encoder maps a short observation context to a representation used only for problem indexing, which selects either a shift-specific local expert or the ID reject option. The selected expert predicts a residual correction $\Delta a_t^{(c)}$, and the final action is $a_t = a_t^{\mathrm{base}} + \Delta a_t^{(c)}$. Thus, JEPA is used for indexing rather than control, while adaptation is achieved through reusable local residual experts.}
\label{fig:pipeline}
\end{figure*}


\section{Introduction}

Visual model-based reinforcement learning (MBRL) has shown strong promise in continuous-control settings by learning compact latent dynamics and planning directly from pixels~\cite{hansen2023td,hafner2023mastering}. However, this promise remains fragile under distribution shift~\cite{levine2020offline}. A policy trained purely on in-distribution (ID) data may perform well in the nominal environment yet fail abruptly when physical parameters change, even when the visual scene appears nearly unchanged. In such cases, the learned world model and planner continue to produce confident action sequences, but those actions are no longer appropriate for the shifted environment~\cite{pinto2017robust,tobin2017domain}.

A natural reaction is to treat this primarily as an OOD detection problem. If the agent can recognize that the current observation or trajectory lies outside the training distribution, perhaps it can penalize risky plans, switch to a safer controller, or directly adapt its policy~\cite{nasvytis2024rethinking}. Indeed, modern joint-embedding models often provide useful latent geometry for this purpose: shifted states may occupy distinct regions of representation space and can often be detected with simple distance-based signals~\cite{maes2026leworldmodel,assran2023self}. However, our experiments reveal a more important point: \emph{detecting shift is not the same as adapting to it}. In visual MBRL, the main challenge is converting shift awareness into better actions.

We arrive at this conclusion through a systematic empirical study. We tested several intuitive strategies for exploiting shift signals, including planning-time penalties, dynamics adaptation, conservative value mixing, direct fine-tuning of the main controller, global residual correction, and coarse or random expert gating. These approaches expose a recurring failure mode. Some methods recognize that the current state is unusual but provide no actionable alternative, so the planner still fails. Others do improve behavior on a target shift but overwrite or destabilize ID performance. Still others produce corrections that are too global, too blunt, or too entangled with the original controller to remain reliable.

These observations motivate a different design principle. Instead of asking a single controller to become universally robust, we keep the strong ID controller intact and grow new knowledge only where it is needed. Concretely, we propose \textbf{JEPA-Indexed Local Expert Growth}. The key idea is to use a frozen Joint Embedding Predictive Architecture (JEPA) only as a \emph{problem indexer}, not as the controller itself. A short observation context is mapped into a JEPA representation, which is then assigned to a problem cluster (or rejected as ID) using a lightweight centroid-based rule. Once a cluster is identified, a corresponding local expert produces a residual action correction on top of the baseline action. The final action is therefore not chosen by replacing the controller, but by augmenting it:
\[
a_t = a^{\text{base}}_t + \Delta a^{(c)}_t,
\]
where \(c\) denotes the selected shift-specific expert, and no correction is applied when the state is rejected as ID.

This modular decomposition is important for two reasons. First, it cleanly separates \emph{recognition} from \emph{adaptation}: JEPA indicates what kind of problem the agent is facing, while the local expert encodes how the action should change. Second, it prevents newly acquired OOD behavior from being written back into the baseline controller, thereby reducing catastrophic forgetting on ID states. In this sense, our method is closer to residual policy learning~\cite{silver2018residual} and modular routing~\cite{yang2020multi} than to multi-controller switching.

We study this framework on DMControl walker-walk under torso-mass shifts. The results show that local expert growth improves OOD return while preserving strong ID performance much better than direct fine-tuning or ungated residual alternatives. Moreover, the learned experts remain effective when the same shift is encountered again under a separate evaluation seed block, supporting a practically meaningful \emph{second-encounter reuse} setting: once the agent has learned how to respond to a recurring shift family, that knowledge can be reused without retraining from scratch.

We also analyze when the method succeeds or fails across different shift families. The results suggest that local expert growth is most effective when the shift induces a consistent baseline failure mode, while high-variance shifts provide a weaker preference signal for residual correction.

Our main contributions are:
\begin{itemize}
    \item We show that OOD detection alone is insufficient for visual MBRL under dynamics shift, because shift signals do not directly specify useful action corrections.
    \item We propose JEPA-Indexed Local Expert Growth, which uses frozen JEPA representations for problem indexing and separate residual experts for local action correction.
    \item We show that problem-specific indexing and separate expert modules are both necessary for preserving ID performance while improving OOD control.
    \item We demonstrate reusable second-encounter adaptation and analyze when the method succeeds or fails across different shift families.
    \item We empirically characterize an asymmetry in visual MBRL shift structure:
automatic ID rejection is reliable with simple density models, whereas fine-grained
OOD sub-family discrimination remains limited by the frozen JEPA representation.
\end{itemize}



\section{Related Work}
\label{sec:related}

\paragraph{Model-Based Reinforcement Learning.}
Model-based reinforcement learning (MBRL) methods learn predictive models of environment dynamics and use them for policy optimization or planning. TD-MPC2~\cite{hansen2023td} combines a visual encoder, latent dynamics model, value function, and policy with latent-space planning, and provides a strong pixel-based baseline for continuous control. Dreamer~\cite{hafner2023mastering} and its successors learn latent world models for actor-critic style policy learning, while MBPO~\cite{janner2019trust} improves sample efficiency by incorporating model-generated rollouts. These methods achieve strong performance in stationary settings, but their behavior under test-time distribution shift remains fragile. Our work builds on this line of research but focuses on a different question: not how to improve the world model itself, but how to preserve strong in-distribution control while attaching targeted corrections under recurring shifts.

\paragraph{Self-Supervised Visual Representation Learning for Control.}
Self-supervised visual learning has become an important ingredient in modern control systems. JEPA~\cite{assran2023self} learn representations by predicting future latent states from past observations and actions, encouraging compact embeddings that capture task-relevant structure. Related representation-learning methods include masked autoencoders (MAE)~\cite{he2022masked}, contrastive methods such as SimCLR~\cite{chen2020simple}, and self-distillation approaches such as DINO~\cite{caron2021emerging}. In the control domain, le-wm~\cite{maes2026leworldmodel} demonstrates that JEPA-style pretraining can support world-model learning from visual trajectories. Our use of JEPA is more limited and more modular: we do not use JEPA as the controller or planner, but rather as a frozen representation whose latent geometry is useful for \emph{problem indexing} under distribution shift.

\paragraph{Robustness and Adaptation Under Distribution Shift.}
A large body of work addresses robustness in reinforcement learning through domain randomization~\cite{tobin2017domain}, robust optimization~\cite{pinto2017robust}, uncertainty-aware control~\cite{chua2018deep,kurutach2018model}, OOD detection~\cite{nasvytis2024rethinking}, and adaptive exploration~\cite{finn2017model}. These approaches typically aim to make a single agent more robust by broadening the training distribution, adding uncertainty penalties, or modifying optimization objectives. Our empirical study suggests a different bottleneck in visual MBRL: detecting shift is often easier than turning that signal into effective new action knowledge. Accordingly, our method does not attempt to train a universally robust controller. Instead, it uses a frozen representation to recognize recurring shift families and attaches separate local experts that learn residual action corrections while leaving the original baseline intact.

\paragraph{Modular Adaptation, Routing, and Residual Correction.}
Modular methods decompose decision-making across multiple models or policies, for example through ensembles~\cite{osband2016deep}, distillation~\cite{rusu2015policy}, soft modular routing~\cite{yang2020multi}, or residual policy learning~\cite{silver2018residual}. These ideas are relevant because our method also separates responsibilities across components. However, our framework differs in two important ways. First, we do not route between two full controllers; the baseline controller always remains active. Second, the learned module is not a replacement policy but a \emph{local residual expert} that adds a shift-specific correction only when the current state is assigned to a known problem cluster. In this sense, our method is closer to modular residual adaptation than to multi-controller switching.

\paragraph{Position of This Work.}
Our work lies at the intersection of representation learning, robustness under shift, and modular adaptation. Relative to prior robust RL methods, we do not broaden the training distribution or inject additional robustness objectives into the main controller. Relative to routing methods, we do not switch between full controllers. Relative to self-supervised world-model methods, we use JEPA not as the primary control model but as a frozen indexer that helps identify recurring shift families. This separation allows representation learning and action adaptation to be evaluated independently.

\section{Method}
\label{sec:method}

\subsection{Problem Setup}
\label{subsec:problem_setup}

We consider visual MBRL under distribution shift. A baseline pixel controller is trained only on in-distribution (ID) data and performs action selection through latent-state planning. At test time, the agent may encounter shifted environments whose physical parameters differ from those seen during training, such as changes in torso-mass. In these settings, the baseline controller often remains strong on ID episodes but becomes brittle under shift.

Our goal is not to replace the baseline controller with a universally robust alternative. Instead, we seek a more targeted form of adaptation: preserve the strong ID behavior of the baseline, detect recurring shift types, and attach \emph{local action corrections} only where they are needed. This leads to a modular formulation in which shift recognition and action adaptation are handled by separate components.

\subsection{Why Detection Alone Is Not Enough}
\label{subsec:why_detection_not_enough}

Distance-based signals in representation space can indicate whether the current state differs from ID data, but they do not define an action-level correction. A large distance from the ID centroid does not specify which action dimension should change, the correction magnitude, or the correction direction. This motivates separating problem indexing from action adaptation.

\subsection{JEPA-Based Problem Indexing}
\label{subsec:jepa_indexing}

We use a frozen JEPA representation only for \emph{problem indexing}, not for control. Given a short observation context
\[
\mathbf{o}_{t-k:t} = (o_{t-k}, o_{t-k+1}, \dots, o_t),
\]
a frozen JEPA encoder produces an embedding
\[
h_t = f_{\mathrm{JEPA}}(\mathbf{o}_{t-k:t}).
\]
This embedding is used to determine whether the current state resembles a known shift family or should be rejected as ID.

\begin{figure}[t]
    \centering
    \begin{subfigure}[t]{0.48\columnwidth}
        \centering
        \includegraphics[width=\linewidth]{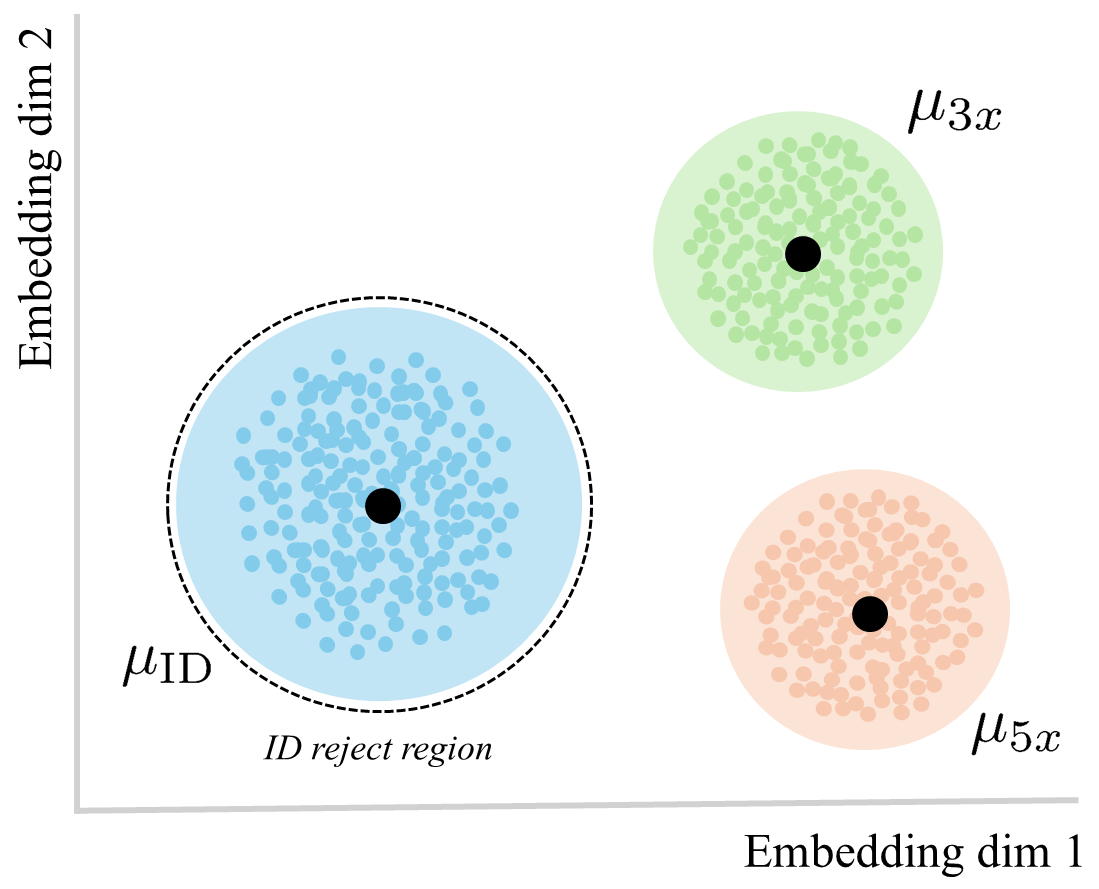}
        \caption{}
    \end{subfigure}
    \hfill
    \begin{subfigure}[t]{0.45\columnwidth}
        \centering
        \includegraphics[width=\linewidth]{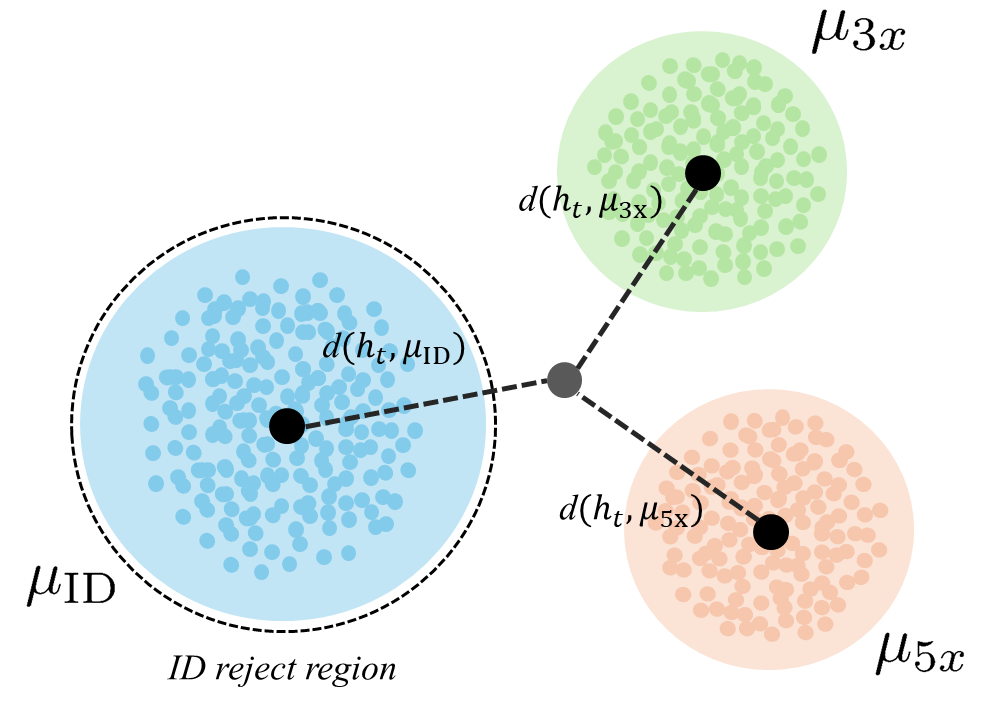}
        \caption{}
    \end{subfigure}
    \vspace{0.5em}

    \begin{subfigure}[t]{0.70\columnwidth}
        \centering
        \includegraphics[width=\linewidth]{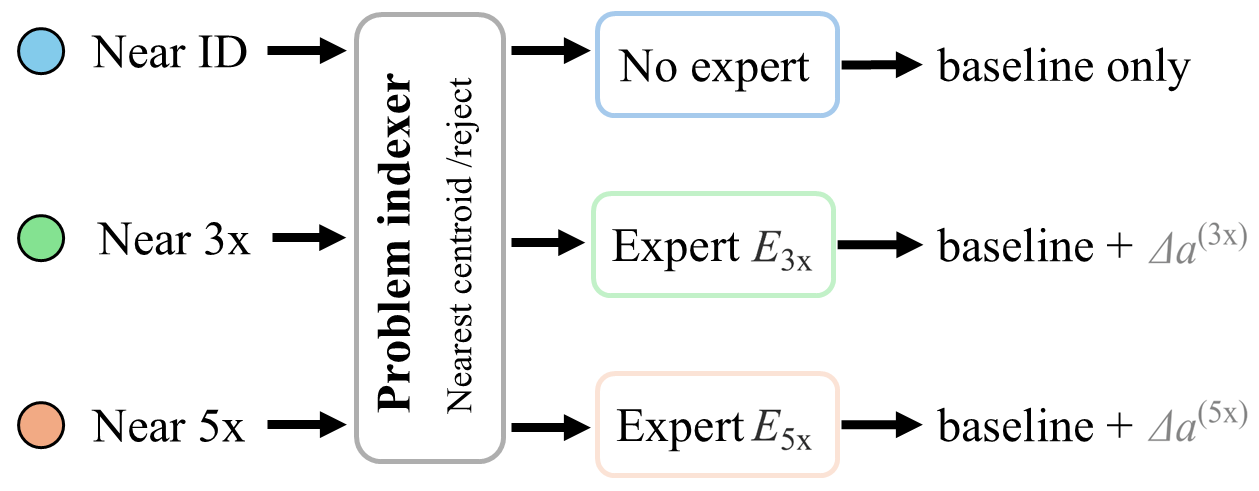}
        \caption{}
    \end{subfigure}

    \caption{A frozen JEPA encoder maps observations to a latent space, where centroid-based assignment selects either a shift-specific expert or the ID reject option.}
    \label{fig:2}
\end{figure}

For each known shift cluster \(c \in \mathcal{C}\), we compute a prototype (centroid) in JEPA space from representative data:
\[
\mu_c = \frac{1}{|\mathcal{D}_c|}\sum_{h \in \mathcal{D}_c} h.
\]
We also maintain an ID centroid \(\mu_{\mathrm{ID}}\). At test time, we assign the current embedding to the nearest centroid,
\[
c^*(h_t) = \arg\min_{c \in \mathcal{C} \cup \{\mathrm{ID}\}} \|h_t - \mu_c\|_2,
\]
subject to an ID reject option. Intuitively, if the embedding is sufficiently close to the ID region, we apply no expert correction; otherwise, we route the state to the most relevant shift-specific expert. Figure~\ref{fig:2} illustrates the indexing mechanism conceptually. The figure illustrates how ID and recurring shift families form separable regions in latent space, how centroid-based assignment selects a local expert, and how the ID reject option prevents unnecessary expert activation. This indexing mechanism plays a deliberately limited role. It does not generate actions and does not modify the baseline controller. Its purpose is only to answer: \emph{what kind of problem does the current state most resemble?}

\subsection{Local Expert Growth}
\label{subsec:local_expert_growth}

Once a state is assigned to a known shift cluster, we apply a corresponding local expert that predicts a residual action correction. Let \(a_t^{\mathrm{base}}\) denote the action produced by the baseline pixel controller. For cluster \(c\), a local expert \(E_c\) outputs
\[
\Delta a_t^{(c)} = E_c(s_t),
\]
where \(s_t\) denotes the controller input state representation used by the expert. The final action is
\[
a_t =
\begin{cases}
a_t^{\mathrm{base}} + \Delta a_t^{(c^*)}, & \text{if } c^*(h_t) \neq \mathrm{ID},\\[4pt]
a_t^{\mathrm{base}}, & \text{if } c^*(h_t) = \mathrm{ID}.
\end{cases}
\]
Thus, the baseline controller always remains active, and the expert only adds a local correction when the current state is assigned to a known problem cluster.

This structure has two important properties. First, it preserves the strong ID behavior of the baseline because no new behavior is written directly into the main controller. Second, it makes the adaptation \emph{problem-specific}: each expert is responsible only for a particular shift family rather than for all OOD states at once.

\subsection{Preference-Based Residual Expert Training}
\label{subsec:expert_training}

A key design choice is that local experts should not simply imitate the baseline action. If an expert is trained to reproduce the baseline, it will learn a near-zero correction and provide no benefit under shift. Instead, the expert must learn how to move the baseline action toward actions associated with better shifted-environment outcomes.

We therefore train each local expert using a pairwise preference objective within its target cluster. From trajectories collected in the target shift environment, we form action pairs consisting of relatively better and worse action instances, denoted \(a^+\) and \(a^-\), according to return-based preference statistics. Starting from the baseline action \(a^{\mathrm{base}}\), the expert predicts a residual \(\Delta a\), and the corrected action
\[
\tilde{a} = a^{\mathrm{base}} + \Delta a
\]
is encouraged to move closer to preferred actions and away from less successful ones.

A generic form of the pairwise objective is
\[
\mathcal{L}_{\mathrm{pref}}
=
\max\Bigl(0,\,
m + d(\tilde{a}, a^+) - d(\tilde{a}, a^-)\Bigr),
\]
where \(d(\cdot,\cdot)\) is an action-space distance and \(m>0\) is a margin. This objective does not force the expert to copy any single action; rather, it learns a correction direction that makes the baseline action more consistent with higher-return behavior in the target cluster.

In practice, this residual-learning formulation is crucial. It allows the expert to encode new action knowledge specific to a recurring shift while remaining local and modular. The HarderPairs variant strengthens this objective by constructing preference pairs from trajectory segments with larger return contrast, rather than using weakly separated action pairs as in the naive-preference variant. 
This makes the supervision signal less ambiguous: the preferred action is more clearly associated with better shifted-environment behavior. 
Harder pair mining therefore changes the training signal, not the inference architecture; the JEPA indexer, ID reject option, and residual expert structure remain unchanged. 
As shown in the experiments, this stronger preference signal is critical for making the local expert stable under 30-seed paired evaluation.
\subsection{ID Reject Option}
\label{subsec:id_reject}

A practical issue in cluster-based assignment is that a state may be closer to one shifted centroid than another even when it is actually ID. Without an explicit reject option, the system may activate an expert unnecessarily and degrade ID performance.

To avoid this, we include an ID centroid and allow the indexer to return \(\mathrm{ID}\) as a valid assignment. When the current embedding is assigned to ID, no expert is applied:
\[
a_t = a_t^{\mathrm{base}}.
\]
This reject mechanism is important for preserving the baseline controller's strong ID performance and for preventing expert over-activation on unrelated or unseen shift types.

\subsection{First Encounter and Second Encounter Reuse}
\label{subsec:reuse}

We evaluate the method in a recurring-shift setting motivated by practical deployment. During a \emph{first encounter}, the agent experiences a target shift family and learns the corresponding local expert. Later, during a \emph{second encounter}, the same shift family reappears under a disjoint evaluation seed block. The learned expert is reused without retraining.

This setting is used to test whether the learned corrections can generalize across disjoint evaluation seeds within the same offset family~\cite{khetarpal2022towards}.


\section{Experiments}
\label{sec:experiments}

\subsection{Experimental Setup}
\label{subsec:exp_setup}

We evaluate on DMControl locomotion tasks under dynamics-related distribution shifts. Our primary benchmark is walker-walk, where the baseline controller is a Hydra 9-channel pixel controller trained on the ID environment with training seed 1. 
Here, ``9-channel'' refers to the stacked visual observation used by the pixel controller, formed by concatenating three RGB frames along the channel dimension. Unless otherwise noted, all main statistical results are reported under a paired-bootstrap protocol with 30 evaluation seeds per condition. We use two disjoint seed blocks: first encounter uses seeds 109000--109029, and second encounter uses seeds 200000--200029. For each seed, we evaluate the baseline and the adapted method on the same environment instance, so that paired deltas can be computed seed-by-seed.

Our primary shifted environments in {walker-walk are \texttt{torso-mass~$3\times$} and \texttt{torso-mass~$5\times$}. These shifts provide a controlled setting in which the baseline remains strong on ID episodes but becomes brittle under recurring dynamics changes. 

We compare:
\begin{itemize}
    \item \textbf{Pixel baseline}: the original ID-trained controller without shift-specific adaptation.
    \item \textbf{V2 (naive preference)}: the earlier local-expert variant trained with the original preference objective.
    \item \textbf{HarderPairs}: the final method, which trains local experts with harder pair mining and is the most stable variant under the 30-seed paired-bootstrap evaluation.
    \item \textbf{Necessity ablations}: direct fine-tuning, global residual correction, coarse gating, and random gating.
\end{itemize}

For the main walker results, we report \textbf{paired delta}, \textbf{95\% bootstrap confidence intervals}, and \textbf{two-sided bootstrap p-values}. For the final method, we also report \textbf{IQM deltas} and \textbf{$P(\mathrm{improve})$}, the probability that the adapted method outperforms the baseline on a randomly sampled evaluation seed. These statistics are intended to distinguish stable improvement from small-sample fluctuations.

We also evaluated the  \textit{cheetah-run} task as a cross-task validation. This experiment tests whether the same indexing-plus-residual-correction design remains effective beyond walker-walk.

For all main results, we report episode returns rather than per-step rewards. This distinction is important because some earlier evaluation summaries used different metrics and should not be mixed with the results reported here.

\subsection{Main Walker Results and Ablations}
\label{subsec:main_results}

Table~\ref{tab:main_results} reports the main paired-bootstrap results under the 30-seed evaluation protocol. Figure~\ref{fig:Per-step cumulative} shows the corresponding mean cumulative return curves across seeds,
illustrating the consistent advantage of HarderPairs on both OOD conditions.
HarderPairs is the only variant that achieves consistent OOD gains across both encounter blocks. 
Figure~\ref{fig:paired_comparison} complements these aggregate statistics by providing a per-seed view of the paired comparison, 
where most points lie above the diagonal, indicating consistent improvement across evaluation seeds. We further analyze two secondary aspects of the main walker results. 
Table~\ref{tab:reuse_delta} evaluates whether the learned correction remains useful on a second encounter with the same shift family. 
Although the second-minus-first differences are not individually significant, the trends show that HarderPairs preserves its OOD correction rather than collapsing after reuse. 
Table~\ref{tab:percentile} examines where the gains occur in the return distribution; the largest improvements appear in lower-percentile episodes, suggesting that the method primarily reduces severe OOD failures rather than uniformly increasing all returns.

\begin{table}[h]
\centering
\caption{\textbf{Main results with paired bootstrap analysis (30 seeds).} Paired delta against pixel baseline, 95\% CI from 10000-resample paired bootstrap with fixed seed (\texttt{RandomState(42)}). V2 (naive preference) shows no consistent OOD improvement under rigorous evaluation; only HarderPairs (harder-pair mining) achieves consistent significance ($p < 0.001$) on all four torso-mass OOD conditions while preserving ID.}
\label{tab:main_results}
\small
\begin{tabular}{lllll}
\toprule
Method & Block & ID & torso-mass~$3\times$ & torso-mass~$5\times$ \\
\midrule
V2 (naive preference) & First  & $+1.31$ \scriptsize{NS}     & $-1.07$ \scriptsize{NS}        & $+0.25$ \scriptsize{NS}        \\
                       & Second & $-4.24$ \scriptsize{NS}     & $-1.43$ \scriptsize{NS}        & $-2.09$ \scriptsize{NEG${}^*$} \\
\midrule
HarderPairs (Ours)     & First  & $+4.68$ \scriptsize{NS}     & $+5.25^{***}$                  & $+5.95^{***}$                  \\
                       & Second & $-3.64$ \scriptsize{NS}     & $+6.91^{***}$                  & $+8.08^{***}$                  \\
\bottomrule
\end{tabular}
\\[2pt]
\scriptsize{${}^*$ $p < 0.05$, ${}^{***}$ $p < 0.001$. NS: not significant. NEG: significant negative delta.}
\end{table}


Direct fine-tuning reduces ID performance by $-20.1$ on the first encounter and $-49.7$ on the second encounter, indicating catastrophic forgetting. Such interference is related to catastrophic forgetting in continual learning~\cite{khetarpal2022towards}.
The global residual expert causes a severe ID drop of $-409.9$, while providing only modest OOD gains.
Coarse gating partially improves OOD performance, but still reduces ID performance by $-40.1$.
Random gating achieves surprisingly competitive OOD results, but it lacks a principled assignment rule.
These results confirm that the failure is not in the OOD gain direction, but in the ID preservation side. This supports the need for both problem-specific indexing and separate expert modules. Detailed IQM and P(improve) statistics are reported in Appendix Table ~\ref{tab:harderpairs_detailed}.


\begin{figure}[t]
    \centering
    \includegraphics[width=0.95\columnwidth]{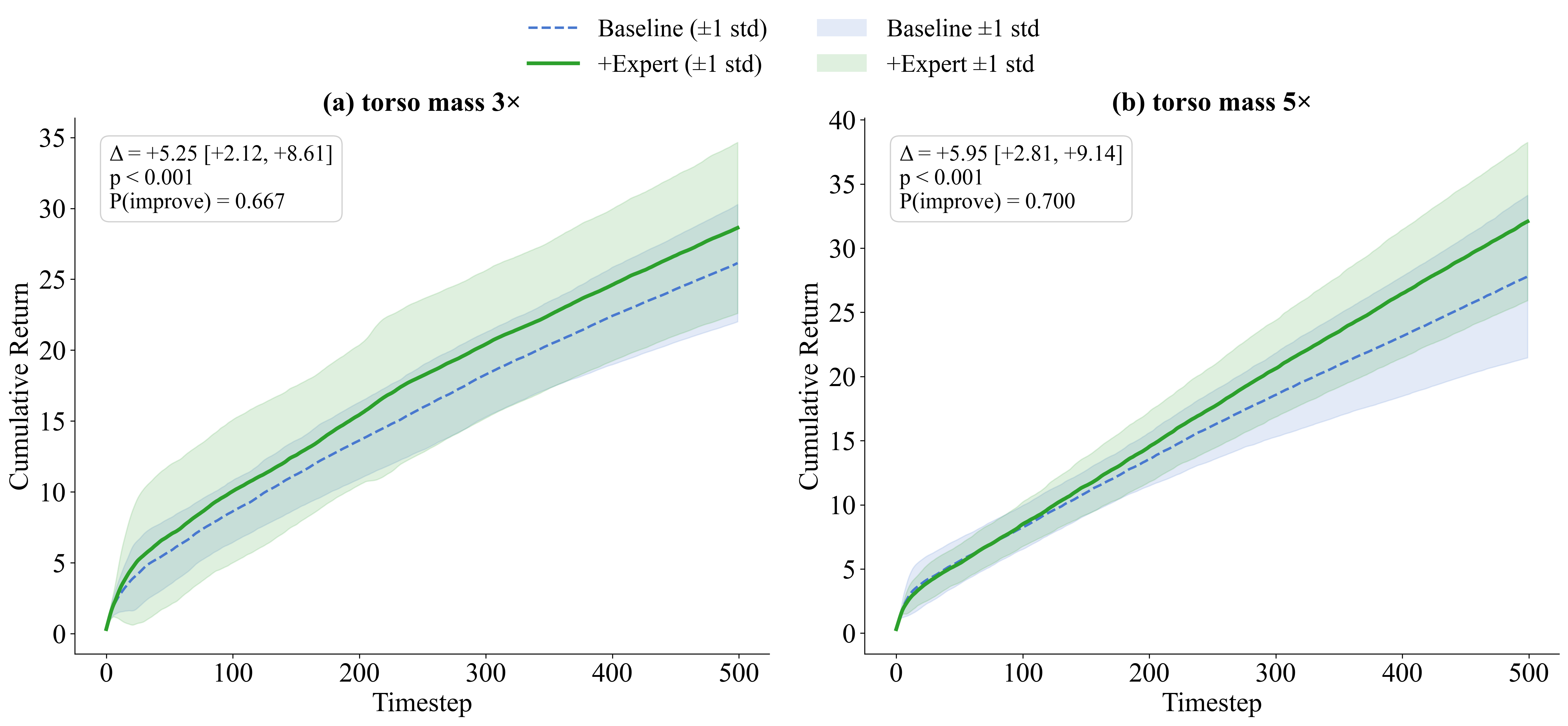}
    \caption{\textbf{Per-step cumulative return on walker-walk torso-mass shifts.}
Mean cumulative return over 30 evaluation seeds for the pixel baseline and HarderPairs.
Shaded bands show $\pm 1$ std. HarderPairs shows a consistent advantage on both torso-mass~$3\times$ and torso-mass~$5\times$.
Statistical results are reported in Table~\ref{tab:main_results}.}
    \label{fig:Per-step cumulative}
\end{figure}

\begin{figure}[t]
    \centering
    \includegraphics[width=\columnwidth]{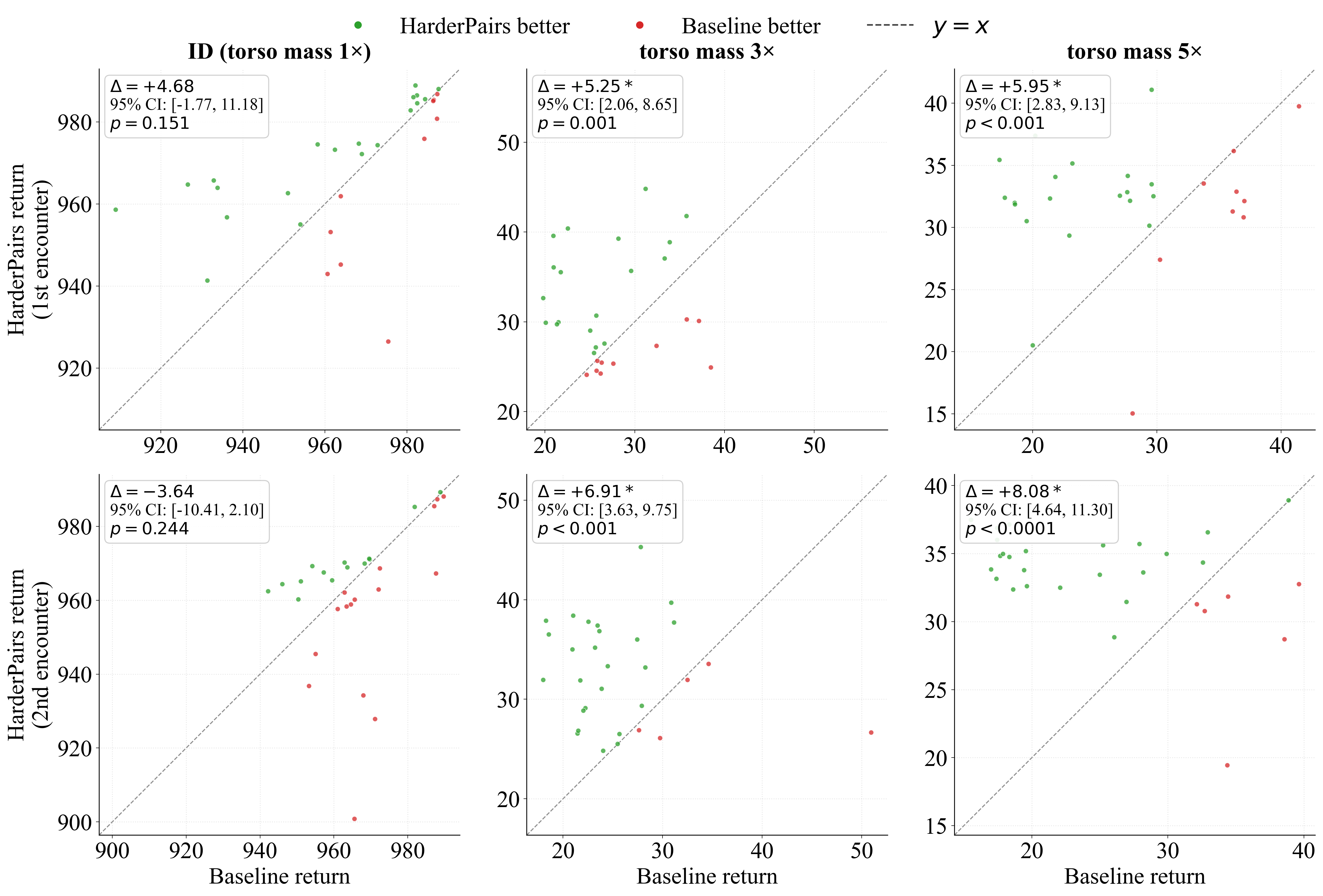}
    \caption{\textbf{Per-seed paired comparison on walker-walk (30 seeds).} 
Each point represents one evaluation seed; coordinates are baseline return
(x) and HarderPairs return (y). Points above the $y=x$ diagonal correspond
to seeds where HarderPairs outperforms baseline. Annotations show paired
$\Delta$, 95\% CI, and $p$-value from 10000-resample paired bootstrap.}
    \label{fig:paired_comparison}
\end{figure}




\begin{table}[t]
\centering
\caption{\textbf{Reuse analysis: second- vs.\ first-encounter paired difference.} For each method and env, we compute \texttt{second\_return$[i]$ - first\_return$[i]$} per seed and report 95\% paired bootstrap CI. No differences are individually significant at $p<0.05$, but HarderPairs exhibits consistent positive trends on both OOD conditions, whereas baseline trends negative. This supports the interpretation that reuse \emph{preserves} (rather than amplifies) the learned correction.}
\label{tab:reuse_delta}
\small
\setlength{\tabcolsep}{5pt}
\begin{tabular}{llccc}
\toprule
\textbf{Method} & \textbf{Env} & \textbf{Second $-$ First} & \textbf{95\% CI} & \textbf{$p$ (2-sided)} \\
\midrule
\multirow{3}{*}{Baseline} 
 & ID  & $+1.63$ & $[-7.34, +11.24]$ & $0.739$ \\
 & 3x  & $-1.41$ & $[-3.80, +1.00]$  & $0.253$ \\
 & 5x  & $-1.32$ & $[-4.68, +1.88]$  & $0.434$ \\
\midrule
\multirow{3}{*}{HarderPairs} 
 & ID  & $-6.69$ & $[-16.28, +2.98]$ & $0.175$ \\
 & 3x  & $+0.24$ & $[-3.02, +3.32]$  & $0.843$ \\
 & 5x  & $+0.81$ & $[-1.39, +3.17]$  & $0.477$ \\
\bottomrule
\end{tabular}
\end{table}

\begin{table}[t]
\centering
\caption{\textbf{Percentile-level improvement of HarderPairs over baseline.} Per-percentile return differences across 30 evaluation seeds. On torso-mass~$5\times$, improvements are concentrated in the lower percentiles ($P_{10}$: $+13.2$, $P_{25}$: $+14.1$ on second encounter), indicating that the method primarily improves worst-case episodes rather than already-successful ones. This tail-robustness pattern is consistent with the core motivation for OOD adaptation in safety-critical deployment.}
\label{tab:percentile}
\small
\setlength{\tabcolsep}{6pt}
\begin{tabular}{llccccc}
\toprule
\textbf{Env} & \textbf{Block} & \textbf{$P_{10}$} & \textbf{$P_{25}$} & \textbf{$P_{50}$} & \textbf{$P_{75}$} & \textbf{$P_{90}$} \\
\midrule
\multirow{2}{*}{torso-mass~$3\times$} 
 & First  & $+3.9$  & $+3.8$  & $+4.2$ & $+6.0$ & $+4.8$ \\
 & Second & $+5.8$  & $+5.6$  & $+8.6$ & $+8.9$ & $+6.7$ \\
\midrule
\multirow{2}{*}{torso-mass~$5\times$} 
 & First  & $+11.3$ & $+11.4$ & $+4.9$ & $+4.8$ & $+3.1$ \\
 & Second & $\mathbf{+13.2}$ & $\mathbf{+14.1}$ & $+8.7$ & $+3.1$ & $+2.1$ \\
\bottomrule
\end{tabular}
\end{table}

\subsection{Cross-Task and Cross-Shift Evaluation}
\label{subsec:cheetah}

We next evaluate cheetah-run as a cross-task validation. The aim is to test whether the same modular design remains useful when the dynamics and gait structure differ from walker-walk. We keep the indexing and local-expert framework unchanged, and use the same paired evaluation protocol where possible.

\begin{figure}[t]
    \centering
    \includegraphics[width=\columnwidth]{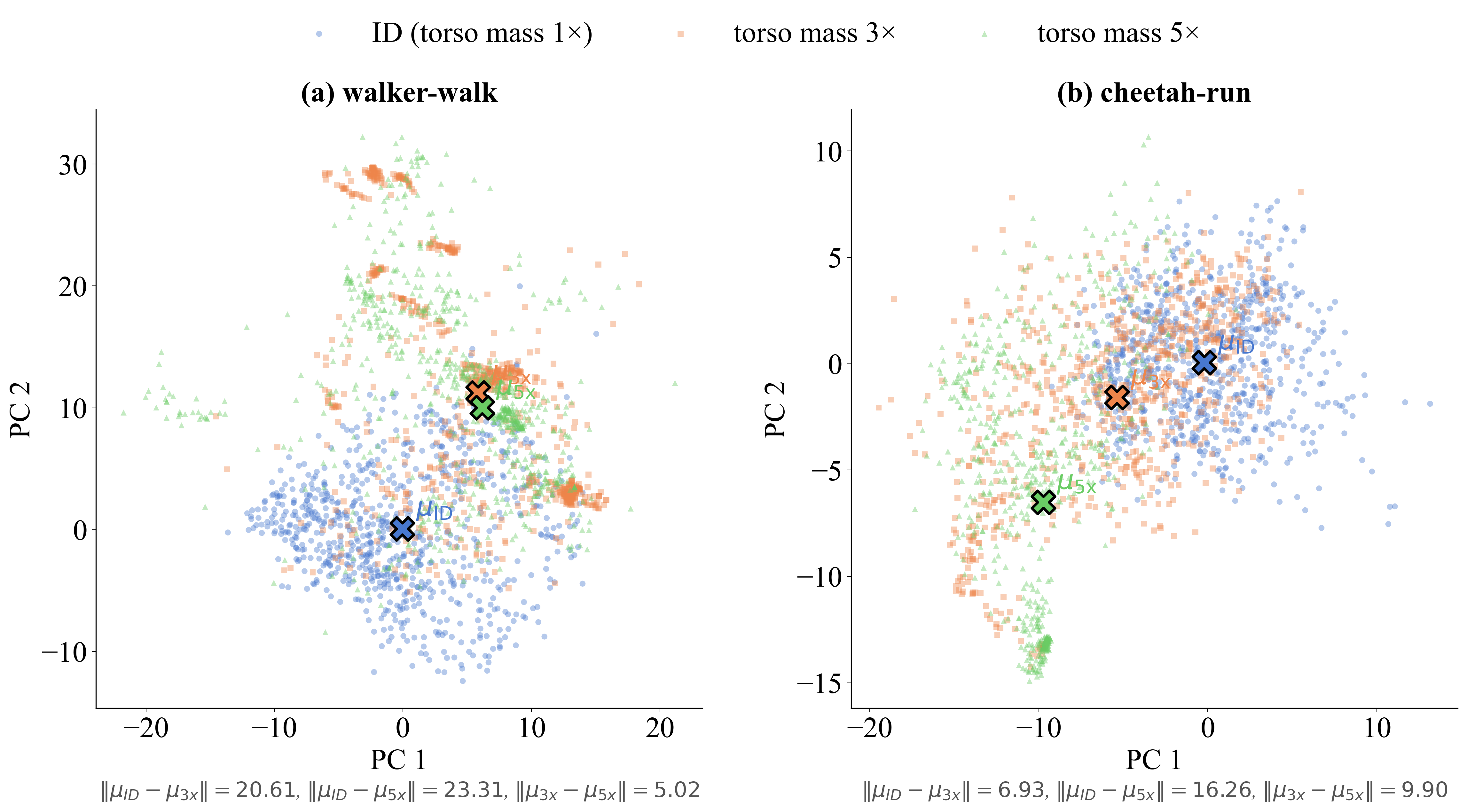}
    
    \caption{\textbf{PCA 2D projection of JEPA embeddings.} 
ID, torso\_mass\_3x, and torso\_mass\_5x states from rollouts of 
\texttt{baseline\_400k.pt}, projected to first two principal components 
of the 64-dim PCA space used by the routing module. 
\textbf{(a) walker-walk}: ID centroid $\mu_{\text{ID}}$ is well-separated 
from shift centroids ($\|\mu_{\text{ID}}-\mu_{3x}\|=20.6$, 
$\|\mu_{\text{ID}}-\mu_{5x}\|=23.3$); however, the two shift centroids 
are close to each other ($\|\mu_{3x}-\mu_{5x}\|=5.0$), explaining 
occasional cross-routing between them. 
\textbf{(b) cheetah-run}: ID and torso-mass~$3\times$ partially overlap 
($\|\mu_{\text{ID}}-\mu_{3x}\|=6.9$), while torso-mass~$5\times$ is 
clearly separated. This motivates the 5x-only routing rule used for 
cheetah-run experiments (Section~\ref{subsec:cheetah}).}
    \label{fig:PCA 2D}
\end{figure}

The cheetah-run results (Table~\ref{tab:cheetah}) support the same qualitative conclusion as the walker study. The baseline again remains strong on ID behavior but degrades under distribution shift, and the harder-pair expert variant improves OOD performance more reliably than the earlier naive-preference version. This is important because cheetah-run has a different control geometry from walker-walk: the gait is faster, the body is more elongated, and the relationship between parameter shifts and visible failure modes is less cleanly separated. In this setting, the success of the harder-pair variant suggests that the benefit is not tied to a single benchmark, but to the way local expert growth converts shift-aware indexing into action correction.

The embedding space corresponding to cheetah-run exhibits lower separability than that of walker-walk, particularly between the ID and torso-mass~$3\times$ settings. This explains why we adopted a torso-mass~$5\times$ routing rule in the cheetah-run setting. Figure~\ref{fig:PCA 2D} visualizes this difference in JEPA embedding geometry.

\begin{table}[h]
\centering
\caption{\textbf{Cross-task validation on cheetah-run (30 seeds, paired bootstrap).} Method uses $5\times$-only routing due to insufficient ID--$3\times$ centroid separability in cheetah JEPA space (ID--$3\times$ distance $7.84$ vs.\ ID--$5\times$ distance $15.43$). Under this routing, torso-mass~$5\times$ shows highly significant improvement ($p<0.001$, $P(\text{improve})=1.0$); torso-mass~$3\times$ shows positive but marginally significant trend; ID is preserved.}
\label{tab:cheetah}
\footnotesize
\begin{tabular}{llrrrr}
\toprule
Env & Block & Paired $\Delta$ & 95\% CI & $p$ (2-sided) & $P(\text{improve})$ \\
\midrule
ID         & First  & $+3.66$  & $[-9.77, +20.04]$ & $0.33$       & $0.67$ \\
           & Second & $+5.98$  & $[-1.99, +14.17]$ & $0.07$       & $0.93$ \\
\midrule
torso\_3x  & First  & $+2.35$  & $[-1.17, +5.77]$  & $0.09$       & $0.91$ \\
           & Second & $+1.05$  & $[-2.11, +4.24]$  & $0.26$       & $0.74$ \\
\midrule
torso\_5x  & First  & $+10.81$ & $[+6.97, +14.52]$ & $<0.001$     & $1.00$ \\
           & Second & $+5.34$  & $[+1.71, +9.14]$  & $0.002$      & $1.00$ \\
\bottomrule
\end{tabular}
\end{table}

We further tested two non-mass-related perturbations for the walker-walk task: \emph{gravity~$5\times$} (gravity scaled by~$5$) and \emph{gear~$0.3$} (actuator gear ratio scaled to~$0.3$). For each perturbation, we trained a dedicated HarderPairs expert model and evaluated it using a dual-category routing rule  (ID vs.\ shift)  under the same paired-bootstrap protocol involving 30 sets of random seeds.

Table~\ref{tab:shift_families} provides a diagnostic analysis of method behavior across different shift families, rather than primary performance results. The gravity expert achieves $\Delta=+2.61$ ($p=0.06$, $P(\text{improve})=0.94$): a positive direction with strong per-seed consistency, marginally not significant under the conservative bootstrap criterion. By contrast, the gear expert produces $\Delta=-17.05$ on its own target environment, despite achieving training preference accuracy~$0.78$ (higher than the torso-mass experts).

This contrast suggests an empirical condition for method success. Effective shifts produce \emph{consistent} baseline failure modes: torso-mass~$3\times$, $5\times$, and gravity~$5\times$ all reduce baseline return to below~$30$ with low variance (Std/Mean ratio~$\le 0.51$). The gear shift, in contrast, exhibits high baseline variance (Std/Mean~$=0.84$): individual gear-environment episodes range from total failure ($\sim$$15$) to near-ID performance ($\sim$$90$). A pairwise preference signal extracted from such trajectories largely reflects episode-to-episode noise rather than learnable corrective directions. Consequently, even a high training preference accuracy does not translate into control improvement.

This finding identifies an empirical boundary for preference-based expert growth: \emph{the method requires shifts that produce consistent baseline degradation}. We further confirmed this boundary by testing alternative perturbations (friction~$\in\{0.05, 0.1, 0.3, 0.5\}$, damping~$\in\{0.5, 2, 5, 10\}$, gravity~$\in\{1.5, 2, 4, 5, 6\}$): only the high-gravity regime ($\geq 4\times$) produces sufficient baseline degradation on walker-walk, and within this regime the variance remains higher than torso-mass shifts, explaining the smaller and only marginally significant gain. Additional supplementary evaluation on unseen shift families is provided in Appendix Table~\ref{tab:supp_400k_reject},
with further analyses of activation reduction and centroid distances in Tables~\ref{tab:supp_old_vs_new} and~\ref{tab:supp_centroid_distance}. This suggests that walker-walk has a relatively narrow set of informative shifts. Future work could formalize this consistency condition as an automatic shift-suitability criterion.





\begin{table}[h]
\centering
\caption{\textbf{Method behavior across shift families on walker-walk.} For each shift, we report the baseline mean return, the baseline-relative degradation, the trained expert preference accuracy, and the paired control $\Delta$ from 30-seed first-encounter evaluation. The method achieves significant improvement when baseline degradation is severe and consistent (torso mass shifts, gravity~$5\times$); it fails when baseline retains moderate performance with high variance (gear~$0.3$). High preference accuracy (gear: 0.78) does not imply control improvement, reinforcing the importance of closed-loop evaluation. \texttt{Std/Mean} of baseline return is reported as a proxy for failure-mode consistency.}
\label{tab:shift_families}
\footnotesize
\begin{tabular}{lrrrrr}
\toprule
Shift & Baseline mean & Std/Mean & Pref.\ acc.\ & Paired $\Delta$ & Significance \\
\midrule
torso-mass~$3\times$ & $26.4$ & $0.21$ & $0.66$ & $+5.25$ & $p < 0.001$ \\
torso-mass~$5\times$ & $26.8$ & $0.22$ & $0.66$ & $+5.95$ & $p < 0.001$ \\
gravity\_$5\times$     & $21.3$ & $0.51$ & $0.78$ & $+2.61$ & marginal ($p=0.06$, $P(\text{imp})=0.94$) \\
gear\_0.3       & $65.2$ & $0.84$ & $0.78$ & $-17.05$ & marginal ($p=0.07$) \\
\bottomrule
\end{tabular}
\\[2pt]
\scriptsize{Selection criterion observed empirically: shifts with \texttt{Std/Mean} $< 0.4$ and severe baseline degradation enable effective expert learning; shifts with high baseline variance do not.}
\end{table}


\begin{figure}[h]
    \centering
    \includegraphics[width=\columnwidth]{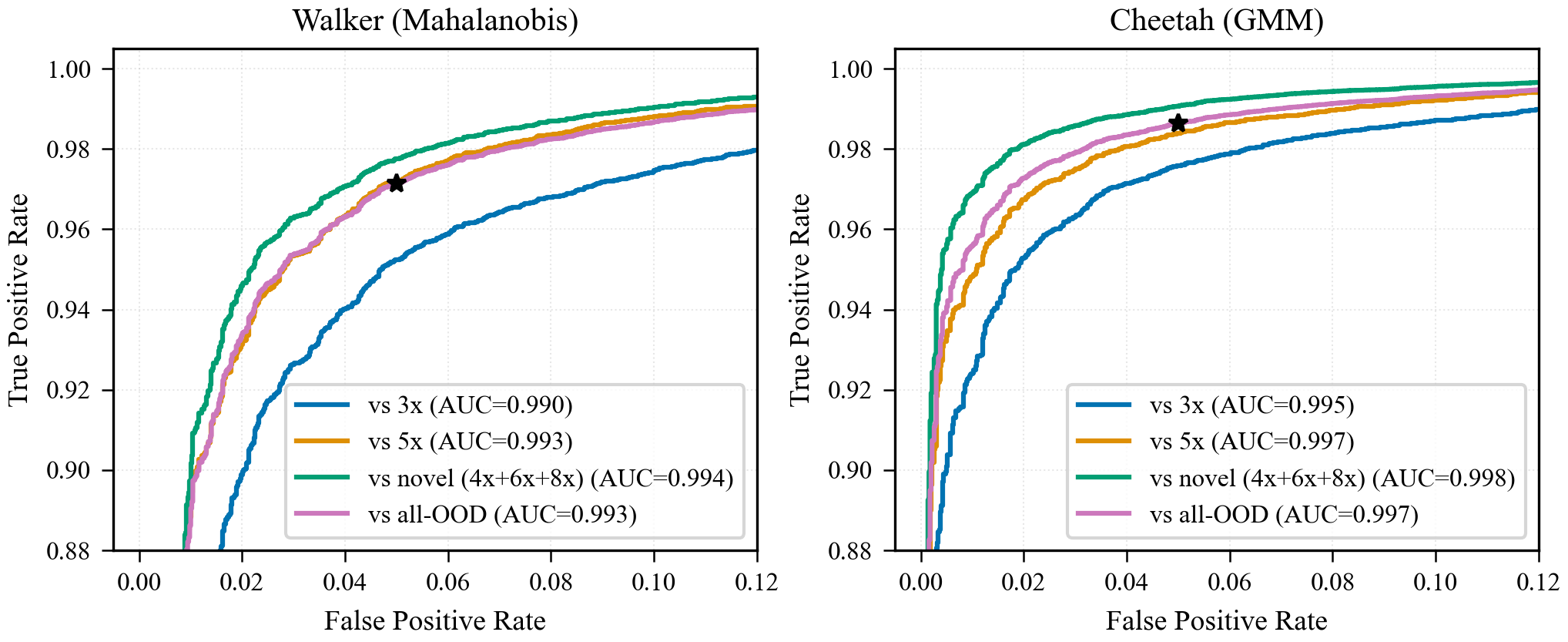}
    \caption{Auto ID-rejection achieves near-perfect discrimination across both tasks. ROC curves for the frozen-JEPA + simple density detector on (a) walker-walk and (b) cheetah-run. Each panel shows discrimination between in-distribution windows and four OOD subsets: known-shift 3$\times$, known-shift 5$\times$, novel-shift family (4$\times$, 6$\times$, 8$\times$ merged), and all OOD combined. The selected operating threshold $\tau$ (5\% in-distribution false-positive rate) is marked with a star. The detector achieves AUC $\geq 0.99$ on both tasks and generalizes to novel shifts unseen during calibration, supporting auto-rejection as a plug-and-play replacement for the manual centroid threshold used in the main routing rule.}
    \label{fig:idreject_roc}
\end{figure}

\begin{figure}[h]
    \centering
    \includegraphics[width=\columnwidth]{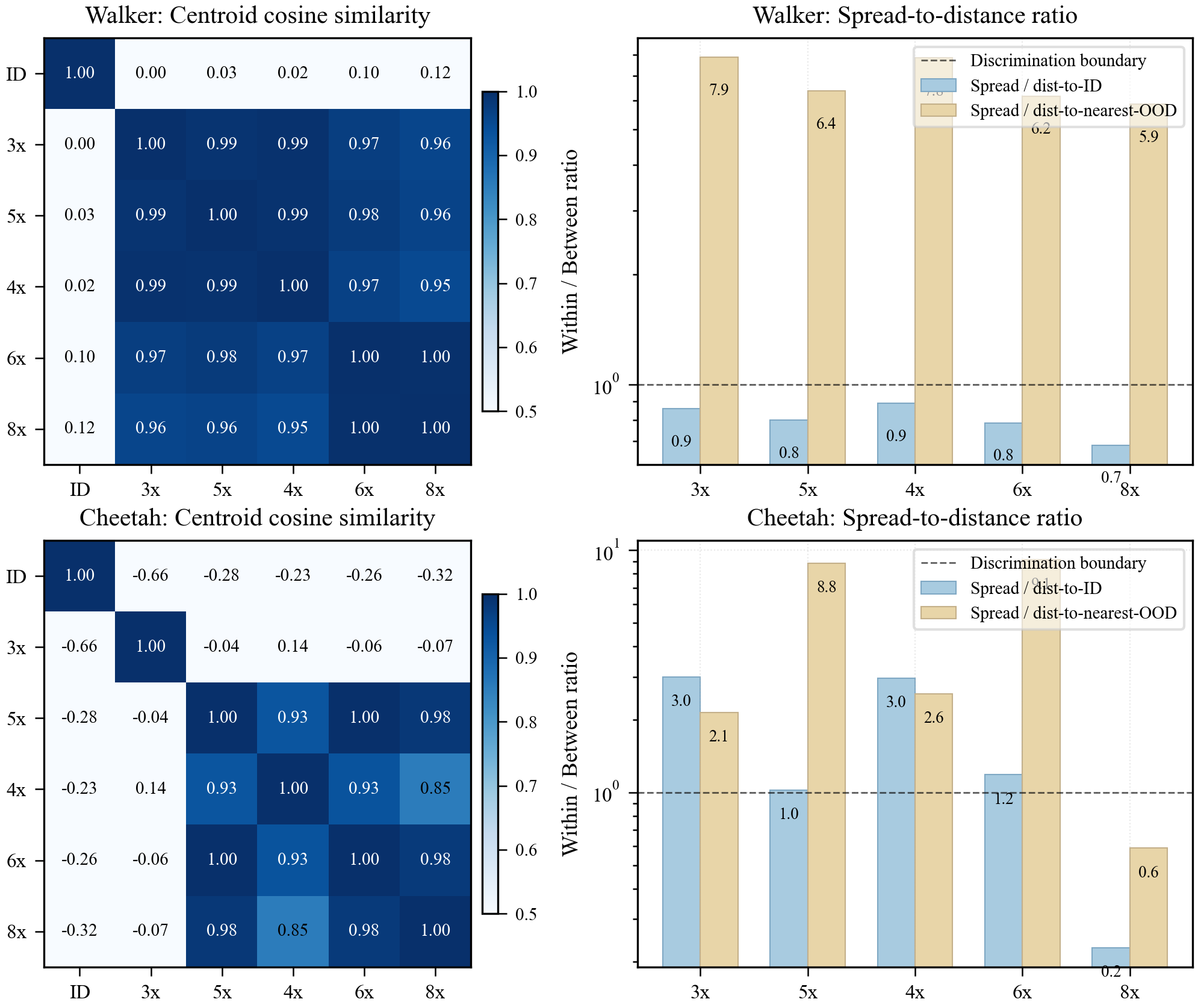}
    \caption{Sub-family discrimination is geometrically infeasible in the frozen-JEPA representation. (a) Pairwise cosine similarity between cluster centroids in the 64-dim PCA space. ID is well-separated from all OOD shifts (similarity $\leq 0.5$), but every pair of OOD shifts is near-collinear (similarity 0.95--0.99) on both tasks. (b) Within-cluster spread relative to between-cluster distance, computed against ID centroid (left bar) and against the nearest OOD centroid (right bar). Ratios above 1$\times$ (dashed line) indicate the discrimination boundary. ID-vs-OOD ratios remain below 1$\times$, while OOD-vs-OOD ratios reach 8.1$\times$ on walker and 4.3$\times$ on cheetah, indicating that OOD subfamilies are not separable along any centroid-based decision rule.}
    \label{fig:C1}
\end{figure}

\section{Discussion}
\label{sec:discussion}

\subsection{Why Detection Alone Is Insufficient}
\label{subsec:discussion_detection}

 The experiments show that representation-space separability does not guarantee useful control adaptation. Detection requires only that a signal correlates with shift; adaptation requires that the signal be converted into a specific, local, correct action change. In our setting, JEPA features reliably satisfy the first condition. The second condition proved harder: a useful detection signal does not automatically specify which dimension of the action should change, by how much, or in what direction. This explains why distance signals alone are insufficient for closed-loop improvement.

The main benefit of the modular design is controlled interference. By assigning detection to JEPA and adaptation to separate local experts, the system never needs a single component to solve both problems simultaneously. The baseline controller is not modified, so ID competence is structurally preserved rather than traded off against OOD gains. The experts are trained only within their target cluster, so the adaptation signal remains local and informative rather than averaged across the full state space.
The 30-seed bootstrap results confirm this. The naive-preference variant, which uses the same architecture but weaker training signal, does not hold up under rigorous evaluation. Therefore, architecture alone is insufficient; under closed-loop evaluation, preference pairs must also remain informative.



\subsection{Limitations}

The current framework has several limitations. First, the learned experts are shift-family specific rather than universally robust. Supplementary experiments show that an expert trained for one shift family does not automatically improve unrelated perturbations, although the ID reject option substantially reduces harmful mis-activation.

Second, the framework assumes that relevant shift clusters are known in advance and does not provide a fully automatic cluster-discovery mechanism, although we report a partial analysis of this issue in Section~\ref{sec:auto_discovery_limits}.

Third, the main analysis is centered on a small number of tasks and recurring shift families, and does not yet establish broad cross-domain generality.

Finally, the second-encounter setting is narrower than full lifelong learning: it demonstrates reuse under repeated shift families, not open-ended adaptation to a continuously changing stream of new environments.

\subsection{On the Limits of Automatic Cluster Discovery}
\label{sec:auto_discovery_limits}

A natural question is whether the manual cluster definitions used in our routing mechanism could be replaced with automatically discovered clusters. We investigated this possibility along three axes: ID-versus-OOD rejection, fine-grained OOD sub-family discrimination, and the implications for routing design. The results reveal a clear asymmetry: ID-versus-OOD detection is achievable with simple density models, while fine-grained discrimination among OOD sub-families is limited by the representation itself.

\paragraph{ID rejection is easy.}
We trained a one-class detector on the frozen JEPA + PCA-64 representation,
fitting a Mahalanobis distance to the ID centroid (walker) and a
$K{=}16$ Gaussian mixture (cheetah). At a calibrated $5\%$ in-distribution
false-positive rate, the detector achieves AUC $\geq 0.99$ on both tasks and
reaches true-positive rates above $95\%$ on every OOD shift tested, including
shifts \emph{not seen during calibration} (torso\_mass $4{\times}$, $6{\times}$,
$8{\times}$). This generalization to unseen shift magnitudes indicates that the
ID region is geometrically compact and well-separated from any non-trivial
dynamics perturbation in the JEPA representation. Figure~\ref{fig:idreject_roc} reports ROC curves and operating points for both tasks. This result establishes that the manual ID-centroid threshold used by the routing rule (Section~\ref{sec:method}) can be replaced by a plug-and-play auto-rejection module without any per-task tuning.

\begin{figure}[t]
    \centering
    \includegraphics[width=\columnwidth]{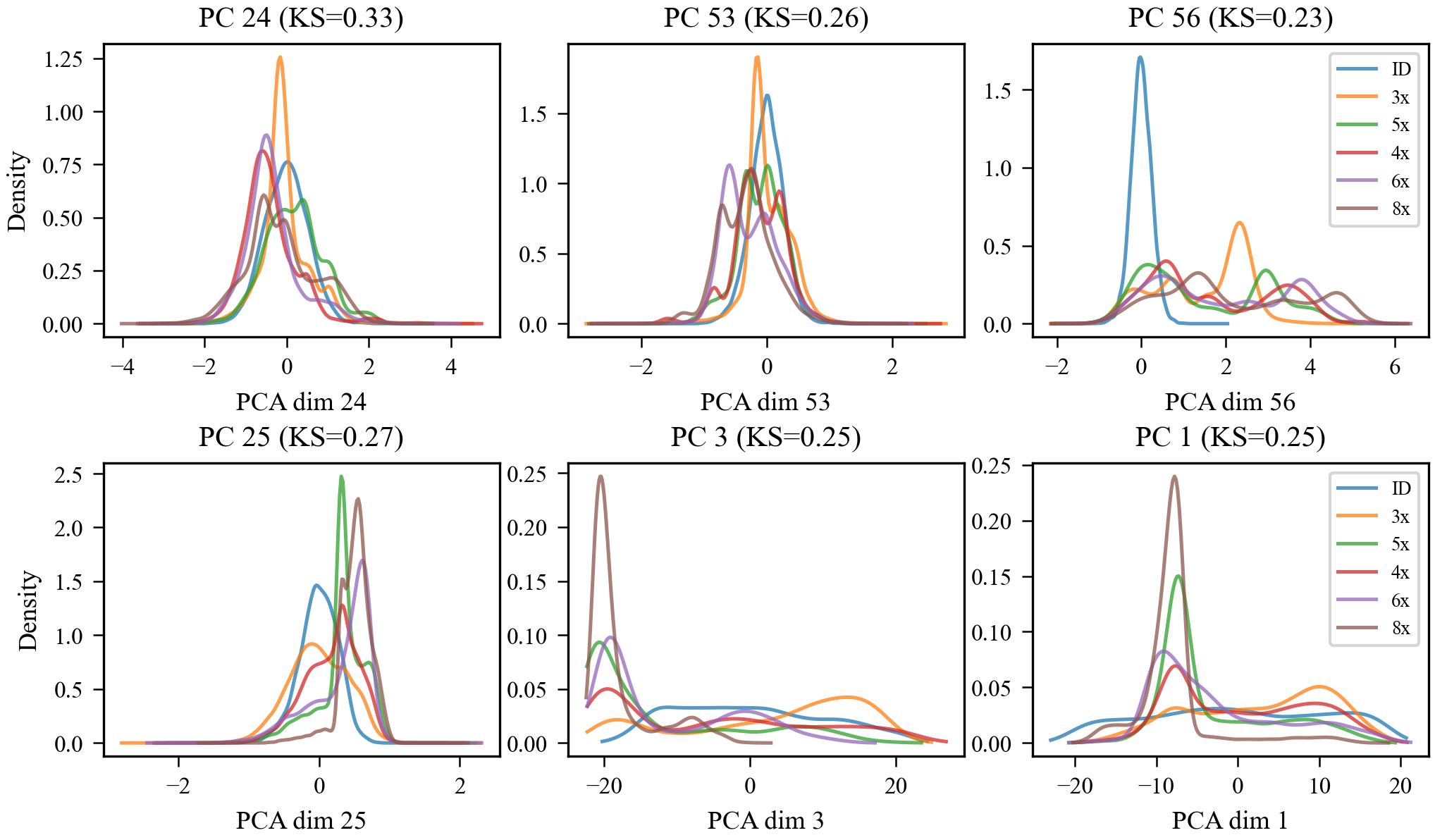}
    \caption{\textbf{OOD subfamilies remain highly overlapping in PCA space.}
Kernel density estimates on the three PCA dimensions with the largest known-vs.-novel OOD KS statistics show that ID is clearly separated, whereas all five OOD shifts largely overlap. This supports the conclusion that OOD sub-family discrimination is representation-limited rather than algorithmically missing.}
    \label{fig:C2}
\end{figure}

\paragraph{Sub-family discrimination is not.}
The same detection geometry that makes ID rejection easy makes OOD sub-family
discrimination essentially infeasible. To diagnose this, we measured pairwise
cluster geometry across six clusters
(ID, $3{\times}$, $5{\times}$, $4{\times}$, $6{\times}$, $8{\times}$) on both
tasks. Figure~\ref{fig:C1} reports two diagnostics. First, the
cosine similarity between OOD centroid pairs lies in
$[0.95, 0.99]$ on both tasks, while the ID centroid is well separated from all
of them. Second, the within-cluster spread exceeds the between-cluster centroid
distance by a factor of $8.1{\times}$ on walker and $4.3{\times}$ on cheetah for
OOD pairs, while the same ratio remains below $1{\times}$ for any
ID-versus-OOD comparison. Figure~\ref{fig:C2} confirms the
same picture at the distributional level: even on the three PCA dimensions
with the highest Kolmogorov--Smirnov statistic, the OOD sub-families heavily
overlap.

To rule out an algorithmic shortcoming, we evaluated eight density-based and
distance-based methods (Joint GMM with $K{\in}\{3,8\}$, per-cluster GMM with
BIC-selected components, Mahalanobis with empirical covariance, $k$-NN with
$k{\in}\{10,50\}$, Isolation Forest, and One-Class SVM). All methods yield
near-chance discrimination between known and novel OOD sub-families
(AUC $\in[0.48, 0.67]$). The systematic failure across both parametric and
non-parametric families indicates that the limitation is representation-level
rather than algorithmic~\cite{winkens2020contrastive,sun2022out}.

\paragraph{Implications for method design.}
This asymmetry has two consequences for our system. First, it justifies the
hybrid design adopted in this paper: automatic ID-rejection combined with
manually defined sub-family routes is not merely a convenient choice but
appears to be \emph{necessary} given current visual representations for MBRL.
Pretrained joint-embedding architectures, despite providing useful latent
geometry for shift detection, do not appear to disentangle physical dynamics
parameters at a granularity sufficient for unsupervised sub-family clustering.
Second, this finding identifies a representation-level bottleneck for future
work on automatic shift discovery in visual MBRL: progress likely requires
representations that explicitly factor dynamics-relevant variation, rather
than scaling existing density-estimation algorithms. We view this as a
complementary failure mode to the high-variance baseline regime documented in
Section~\ref{subsec:cheetah} (e.g., gear $0.3$ on walker), and together these
two boundaries delineate the operational scope of preference-based local
expert growth.




\section{Conclusion}

This paper studied modular adaptation for visual MBRL under dynamics shift. We proposed JEPA-Indexed Local Expert Growth, which uses frozen JEPA features for problem indexing and local residual experts for action correction. Under paired-bootstrap evaluation, HarderPairs improved torso-mass OOD returns while keeping ID changes non-significant.

The results support two conclusions. First, in visual MBRL the main bottleneck under distribution shift is not recognizing that a state is unusual, but attaching the right local action correction once that shift has been recognized. Second, modular adaptation is a more stable strategy than monolithic retraining in this setting. The learned experts also remained reusable on second encounter, supporting the view that adaptation can be framed as incremental knowledge growth rather than repeated end-to-end reoptimization. We additionally found that automatic ID rejection generalizes reliably 
(AUC $\geq 0.99$) including to unseen shift magnitudes, while fine-grained 
OOD sub-family discrimination remains representation-limited under all eight 
density-based methods we evaluated. This asymmetry justifies our hybrid 
manual-routing + auto-reject design and identifies disentangled dynamics 
representation as a key bottleneck for future automatic shift-discovery work.

We further evaluated the method on a second perturbation family (gravity 5x) and on a second control task (cheetah-run). Both confirmed the same qualitative pattern: when the shift produces a sufficiently consistent baseline failure mode, local expert growth improves OOD return without damaging ID. We also identified a regime where the method does not work: shift families with high baseline variance, such as gear ratio changes on walker-walk, do not provide a stable preference signal, and a corresponding expert fails to learn a useful correction. This delineates a clear empirical boundary for the method.

Overall, the results support a modular design principle: preserve the ID controller, use representation geometry to identify recurring shift families, and store new action knowledge in reusable local experts.



\clearpage
\appendix

\section{Additional Results}

\begin{table}[h]
\centering
\caption{Supplementary evaluation on unseen shift families using the 400K backbone with the ID reject option enabled. The experts are still trained only for torso-mass~$3\times$ and torso-mass~$5\times$; no new experts are trained for these supplementary shifts.}
\label{tab:supp_400k_reject}
\resizebox{\textwidth}{!}{
\begin{tabular}{lcccccccc}
\hline
Shift & Baseline & Expert & $\Delta$ & Reject-to-ID & Assign-to-3x & Assign-to-5x & Total Activation & Conclusion \\
\hline
torso-mass~$2\times$ & $40.1 \pm 16.0$ & $35.9 \pm 17.6$ & $-4.2$ & 29.8\% & 50.6\% & 19.6\% & 70.2\% & Partial same-family transfer \\
gear\_0.3       & $62.5 \pm 34.1$ & $62.2 \pm 45.8$ & $-0.3$ & 45.7\% & 54.3\% & 0.0\%  & 54.3\% & Safe fallback (near-zero harm) \\
friction\_0.2   & $934.9 \pm 31.3$ & $926.9 \pm 28.3$ & $-8.1$ & 98.9\% & 1.1\%  & 0.0\%  & 1.1\%  & Safe fallback (near-complete reject) \\
damping\_2x     & $975.5 \pm 10.7$ & $972.8 \pm 14.7$ & $-2.7$ & 99.6\% & 0.4\%  & 0.0\%  & 0.4\%  & Safe fallback (near-complete reject) \\
\hline
\end{tabular}
}
\end{table}

\begin{table}[h]
\centering
\caption{Comparison between the old supplementary evaluation (without ID reject option) and the new 400K supplementary evaluation with ID reject option. The reject option substantially reduces unnecessary expert activation and improves safety on unseen shift families.}
\label{tab:supp_old_vs_new}
\resizebox{\textwidth}{!}{
\begin{tabular}{lccccc}
\hline
Shift & Old Activation & New Activation & Old $\Delta$ & New $\Delta$ & Improvement \\
\hline
torso-mass~$2\times$ & 100\% & 70.2\% & $-3.8$  & $-4.2$ & $-30\%$ activation, slightly worse $\Delta$ \\
gear\_0.3       & 100\% & 54.3\% & $-2.8$  & $-0.3$ & $-46\%$ activation, $\sim 15\times$ smaller harm \\
friction\_0.2   & 100\% & 1.1\%  & $-23.2$ & $-8.1$ & $\sim 99\%$ activation reduction, $\sim 3\times$ smaller harm \\
damping\_2x     & 100\% & 0.4\%  & $-18.0$ & $-2.7$ & $\sim 99\%$ activation reduction, $\sim 7\times$ smaller harm \\
\hline
\end{tabular}
}
\end{table}

\begin{table}[h]
\centering
\small
\caption{Average distance from unseen-shift embeddings to the ID and shift centroids. torso-mass~$2\times$ is geometrically closest to the 3x centroid, suggesting partial same-family transfer in JEPA space.}
\label{tab:supp_centroid_distance}
\begin{tabular}{lcccc}
\hline
Shift & dist$_{\mathrm{ID}}$ & dist$_{3x}$ & dist$_{5x}$ & Nearest centroid \\
\hline
torso-mass~$2\times$ & 18.71 & 12.57 & 14.52 & 3x \\
gear\_0.3       & 16.65 & 16.66 & 19.82 & borderline (ID / 3x) \\
friction\_0.2   & 10.02 & 20.50 & 23.87 & ID \\
damping\_2x     & 9.79  & 20.50 & 23.86 & ID \\
\hline
\end{tabular}
\end{table}

\begin{table}[H]
\centering
\caption{\textbf{Detailed HarderPairs paired-bootstrap statistics on walker-walk (30 seeds).} Paired $\Delta$, 95\% CI, and $p$-value computed via 10000-resample paired bootstrap. IQM denotes interquartile mean. $P(\text{improve})$ is the per-seed probability that the method outperforms the baseline.}
\label{tab:harderpairs_detailed}
\footnotesize
\resizebox{\textwidth}{!}{%
\begin{tabular}{llcccccc}
\toprule
Env & Block & Paired $\Delta$ & 95\% CI & $p$ & IQM $\Delta$ & IQM CI & $P(\text{imp.})$ \\
\midrule
ID & First  & $+4.68$ & $[-1.77, +11.18]$ & $0.151$ & $+3.20$ & $[-0.36, +8.19]$ & $0.667$ \\
ID & Second & $-3.64$ & $[-10.43, +2.28]$ & $0.247$ & $-0.48$ & $[-4.91, +3.22]$ & $0.467$ \\
\midrule
torso-mass~$3\times$ & First  & $+5.25$ & $[+2.12, +8.61]$ & $0.0006$ & $+4.35$ & $[+1.27, +8.17]$ & $0.667$ \\
torso-mass~$3\times$ & Second & $+6.91$ & $[+3.62, +9.87]$ & $<0.0001$ & $+7.55$ & $[+4.33, +10.60]$ & $0.833$ \\
\midrule
torso-mass~$5\times$ & First  & $+5.95$ & $[+2.81, +9.14]$ & $0.0002$ & $+5.83$ & $[+1.92, +9.58]$ & $0.700$ \\
torso-mass~$5\times$ & Second & $+8.08$ & $[+4.65, +11.36]$ & $<0.0001$ & $+9.33$ & $[+4.88, +13.24]$ & $0.800$ \\
\bottomrule
\end{tabular}%
}
\end{table}


\bibliography{example}  

\end{document}